\documentclass[11pt]{article}

\usepackage[preprint]{acl}

\usepackage{times}
\usepackage{latexsym}

\usepackage[T1]{fontenc}

\usepackage[utf8]{inputenc}

\usepackage{microtype}

\usepackage{inconsolata}

\usepackage{graphicx}
\definecolor{cvprblue}{rgb}{0.21,0.49,0.74}

\usepackage{epigraph} 
\usepackage{wrapfig}
\usepackage{url}
\usepackage{amsmath}  
\usepackage{amssymb}  
\usepackage{amsfonts} 

\usepackage{algorithm}
\usepackage{algorithmic}

\usepackage{bbding}
\usepackage{pifont}
\usepackage{wasysym}

\usepackage{verbatim}
\usepackage{booktabs}
\usepackage{multirow}
\usepackage{colortbl}
\usepackage{adjustbox}
\usepackage{makecell}
\usepackage{xspace}
\usepackage{mathtools}
\usepackage{graphicx}
\usepackage{enumitem}
\usepackage{textcomp}
\usepackage{pifont}
\usepackage{bm}

\usepackage{cleveref} 

\usepackage[table]{xcolor}

\title{ Revealing and Enhancing Core Visual Regions:  Harnessing Internal Attention Dynamics for Hallucination Mitigation in LVLMs }

\title{When Attention Dynamics Matter: Revealing and Enhancing Core Visual Regions for Hallucination Mitigation in LVLMs}

\author{
  \textbf{Guangtao Lyu\textsuperscript{1}},
  \textbf{Qi Liu\textsuperscript{1}},
  \textbf{Chenghao Xu\textsuperscript{3}},
  \textbf{Jiexi Yan\textsuperscript{2}},
  \textbf{Muli Yang\textsuperscript{4}},
  \textbf{Xueting Li\textsuperscript{1}},
  \textbf{Fen Fang\textsuperscript{4}},
  \textbf{Cheng Deng\textsuperscript{1}} \\
  \\
  \textsuperscript{1}School of Electronic Engineering, Xidian University, Xi'an, China \\
  \textsuperscript{2}School of Computer Science and Technology, Xidian University, Xi'an, China \\
  \textsuperscript{3}College of Computer and Information, Hohai University, Nanjing, China \\
  \textsuperscript{4}Institute for Infocomm Research, A*STAR, Singapore \\
  \small{
    \textbf{Correspondence:} Cheng Deng \textless chdeng.xd@gmail.com\textgreater
  }
}

\begin{document}
\maketitle
\begin{abstract}

LVLMs have achieved strong multimodal reasoning capabilities but remain prone to hallucinations, producing outputs inconsistent with visual inputs or user instructions. Existing training-free methods, including contrastive decoding and auxiliary expert models, which incur several times more computational overhead and may introduce potential interference, as well as static internal signal enhancement, are often vulnerable to the attention sink phenomenon.
We find that internal Positive Attention Dynamics (PAD) in LVLMs naturally reveal semantically core visual regions under the distortions of attention sinks. Based on this, we propose Positive Attention Dynamics Enhancement (PADE), a training-free attention intervention that constructs a PAD map to identify semantically core visual regions, applies per-head Median Absolute Deviation Scaling to adaptively control the intervention strength, and leverages System-Token Compensation to maintain attention to complex user instructions and support long-term output consistency.
Experiments on multiple LVLMs and benchmarks show that PADE improves visual grounding and reduces hallucinations, validating the effectiveness of leveraging internal attention dynamics for reliable multimodal reasoning.

\end{abstract}

\begin{figure}[t]
    \centering
    \includegraphics[width=1\linewidth]{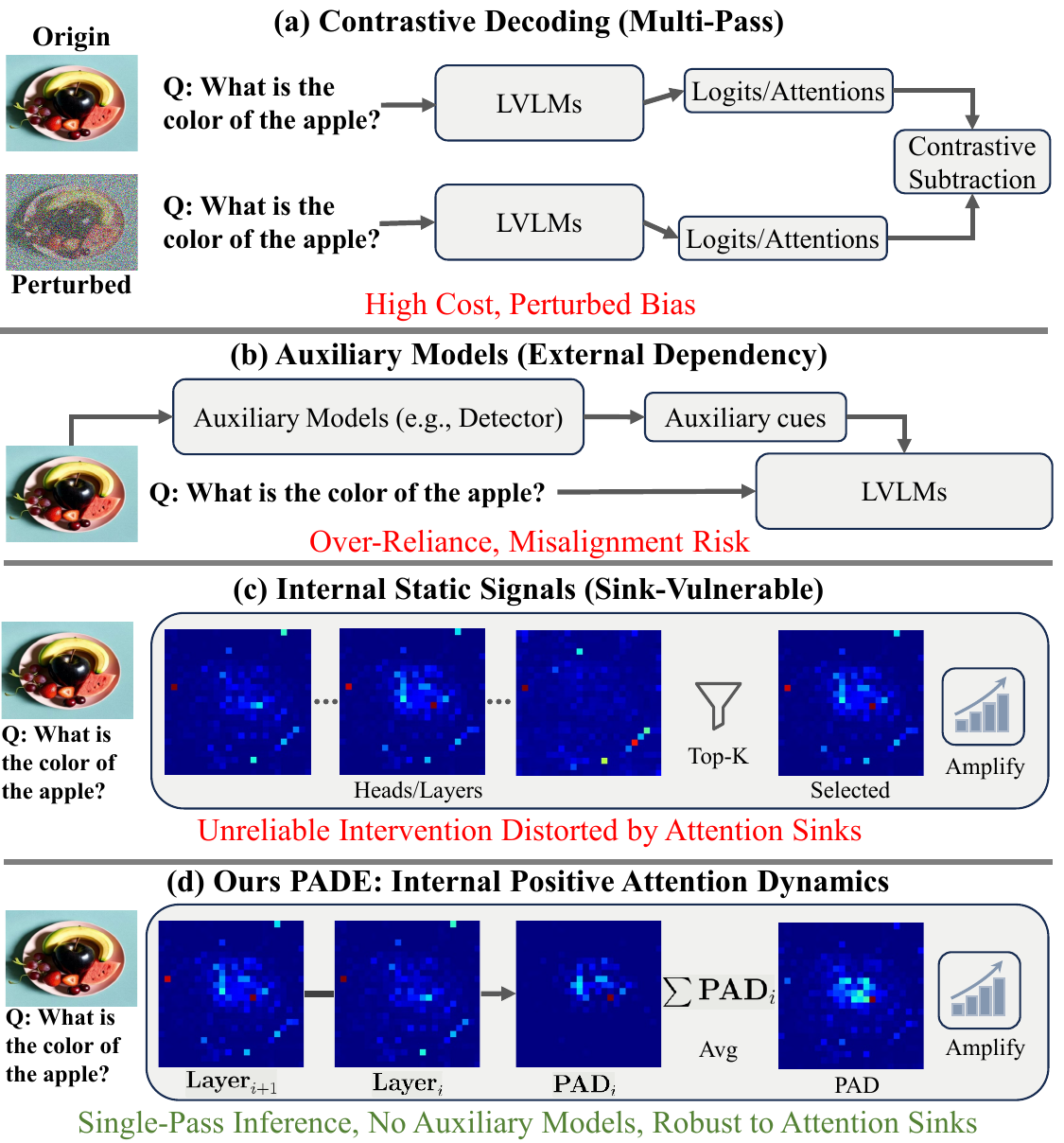}
\caption{Comparison of hallucination mitigation paradigms.
(a) Contrastive decoding methods.
(b) Auxiliary expert methods.
(c) Static internal signal methods.
(d) Ours PADE:  internal positive attention dynamics.}
    \label{fig:intro_pipeline_compare}
\end{figure}

\section{Introduction}\label{sec:intro}

Large Vision Language Models (LVLMs) have achieved remarkable progress in multimodal reasoning and demonstrate strong performance across a wide range of tasks~\cite{llava,gpt4,qwenvl}. Despite these advances, LVLMs remain prone to hallucinations~\cite{liu2024survey_lvlm,ji2023_hallucination_nlp_survey}, generating content inconsistent with visual inputs or user instructions, which undermines their reliability, particularly in safety-critical applications such as medical analysis~\cite{sun2024self_medical,chen2024towards_medical,hu2024omnimedvqa_medical} and autonomous driving~\cite{jiang2024senna_driving,sun2025towards_driving,shao2024lmdrive_driving}.

Recent studies have shown that hallucinations in LVLMs primarily arise from an over-reliance on linguistic priors and insufficient utilization of visual inputs~\cite{bai2024hallucination_survey_llm,ji2023_hallucination_nlp_survey,liu2024survey_lvlm}. To address this issue, many inference-time intervention methods have been proposed to enhance visual grounding by amplifying vision-related signals, such as visual tokens, attention activations, or output logits.
Existing strategies mainly fall into three types (\Cref{fig:intro_pipeline_compare}):
(1) \textbf{contrastive decoding}, which amplifies visual contributions by contrasting outputs generated under different visual conditions (e.g., PAI~\cite{pai_lack_visual}, IBD~\cite{zhu2024ibd}, VCD~\cite{vcd_lack_visual}). These methods require multiple forward passes and may introduce additional bias from the contrastive perturbed signal.
(2) \textbf{auxiliary expert models}, which leverage external models to provide auxiliary cues or highlight salient regions (e.g., HALC~\cite{halc_lack_visual}, AGLA~\cite{an2025mitigating_agla__lack_visual}), at the cost of introducing external dependencies and potential semantic misalignment with the target LVLM.
and (3) \textbf{internal static signals}, which enhance vision-related heads, layers, or tokens by selecting top-ranked elements based on attention values or other heuristic scores (e.g., VHR~\cite{he2025cracking_vhr}, VAF~\cite{yin2025clearsight_vaf}).
Relying on static criteria such as top-$k$ selection or thresholding, these methods are highly vulnerable to the {attention sink phenomenon}~\cite{xiao2023efficient_attention_sinks_ori,kang2025see_attention_sinks_var}, where dominant but semantically irrelevant sink tokens are repeatedly amplified, biasing attention away from truly informative visual regions.

To address these limitations, we revisit a key question:
\emph{how can semantically core visual regions be reliably identified and enhanced
in the presence of attention sink distortions,
without relying on external models or inputs?}
Our key finding is that {positive attention dynamics across layers reveal semantically core visual regions (\Cref{fig:motivation_pad_vs_static}).
Core regions exhibit stronger positive inter-layer attention changes, 
while irrelevant regions remain weakly attended, and attention sinks show irregular fluctuations.
Leveraging these internal {Positive Attention Dynamics (PAD)} allows us to {reveal visual evidence that emerges coherently through the model’s internal understanding process,
thereby enabling more reliable identification of semantically core visual regions without external auxiliary models.

Motivated by this observation, we propose Positive Attention Dynamics Enhancement (PADE), a training-free attention intervention that selectively reinforces semantically core visual regions to improve visual grounding and mitigate hallucinations. PADE identifies core regions by leveraging the model’s internal {positive attention dynamics} and enhances them in the target layer, without relying on external models or multiple forward passes.
Specifically, PADE constructs a {Positive Attention Dynamics (PAD)} map from positive inter-layer attention deltas to highlight semantically core regions. To adaptively control the intervention strength, PAD is scaled per attention head using the {Median Absolute Deviation (MAD)}, ensuring robustness to extreme values while preserving proportionality to the underlying signal. Finally, we propose {System-Token Compensation (STC)}, which leverages system tokens with high attention ratios but limited semantic relevance as a compensation source, thereby preserving attention to complex instructions and maintaining long-term generation.

In summary, our contributions are as follows:
\begin{itemize}
    \item We demonstrate that internal Positive Attention Dynamics (PAD) provide a more reliable signal for identifying semantically core visual regions than static signal-based metrics, especially under the distortions of attention sinks.
    \item We propose {Positive Attention Dynamics Enhancement (PADE)}, a training-free attention intervention that leverages PAD to identify and selectively reinforce semantically core visual regions during inference.
    \item Extensive experiments on both hallucination-focused and general-purpose benchmarks show that PADE effectively improves visual grounding and reduces hallucinations, while preserving overall multimodal understanding.
\end{itemize}

\begin{figure}[t]
    \centering
    \includegraphics[width=0.99\linewidth]{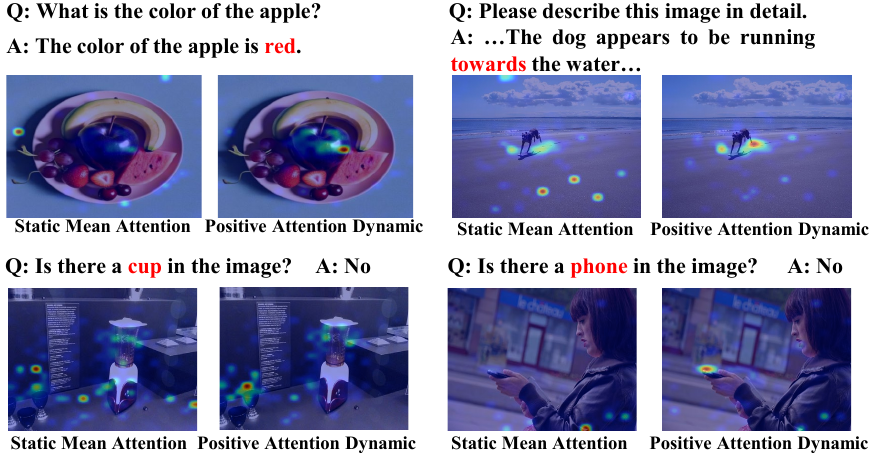}
\caption{
Static versus dynamic internal attention signals.
Static mean attention is dominated by attention sinks, while Positive Attention Dynamics (PAD) more reliably highlight semantically core visual regions.
}
    \label{fig:motivation_pad_vs_static}
\end{figure}

\begin{figure*}[t]
    \begin{minipage}{0.49\textwidth}
        \centering
    \includegraphics[width=0.99\linewidth]{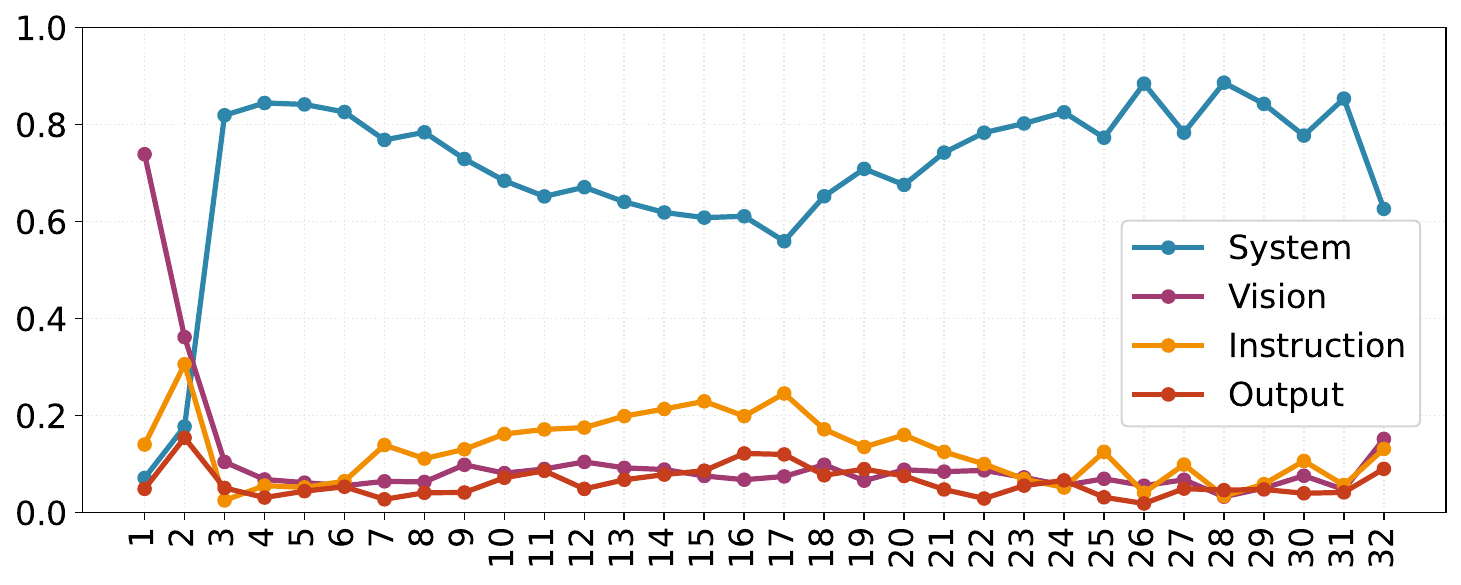}
    \end{minipage}
    \begin{minipage}{0.49\textwidth}
        \centering
    \includegraphics[width=0.99\linewidth]{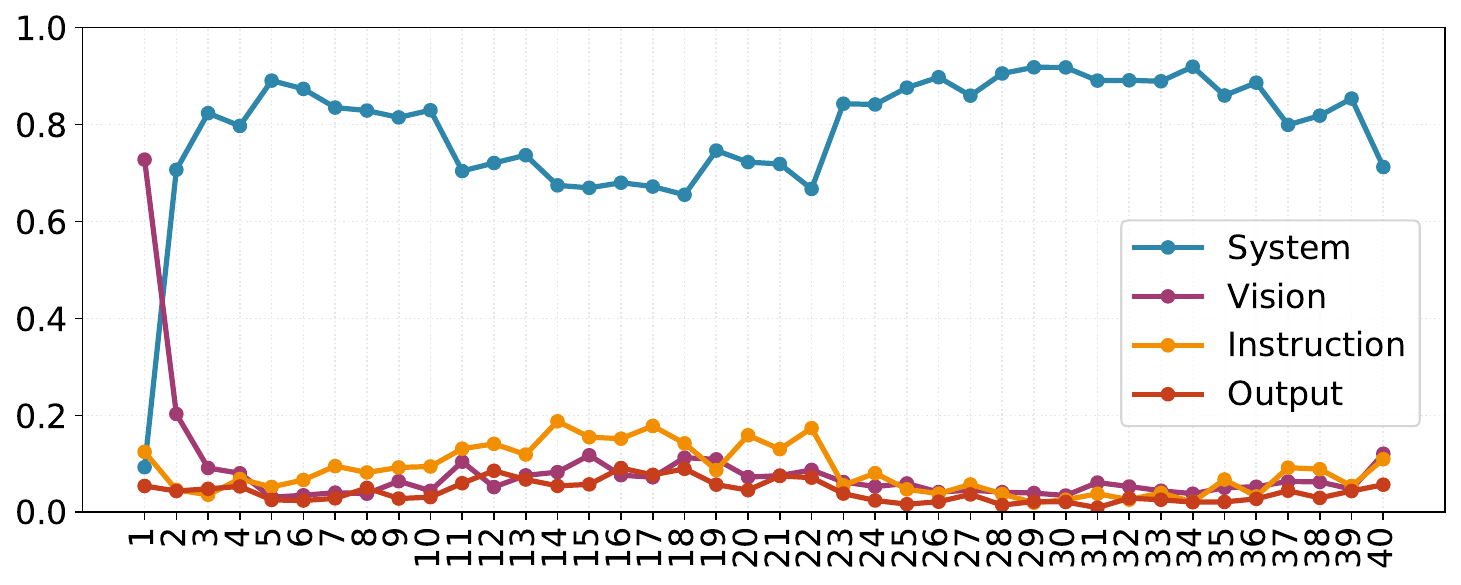}
    \end{minipage}
    
    \begin{minipage}{0.49\textwidth}
        \centering
    \includegraphics[width=0.99\linewidth]{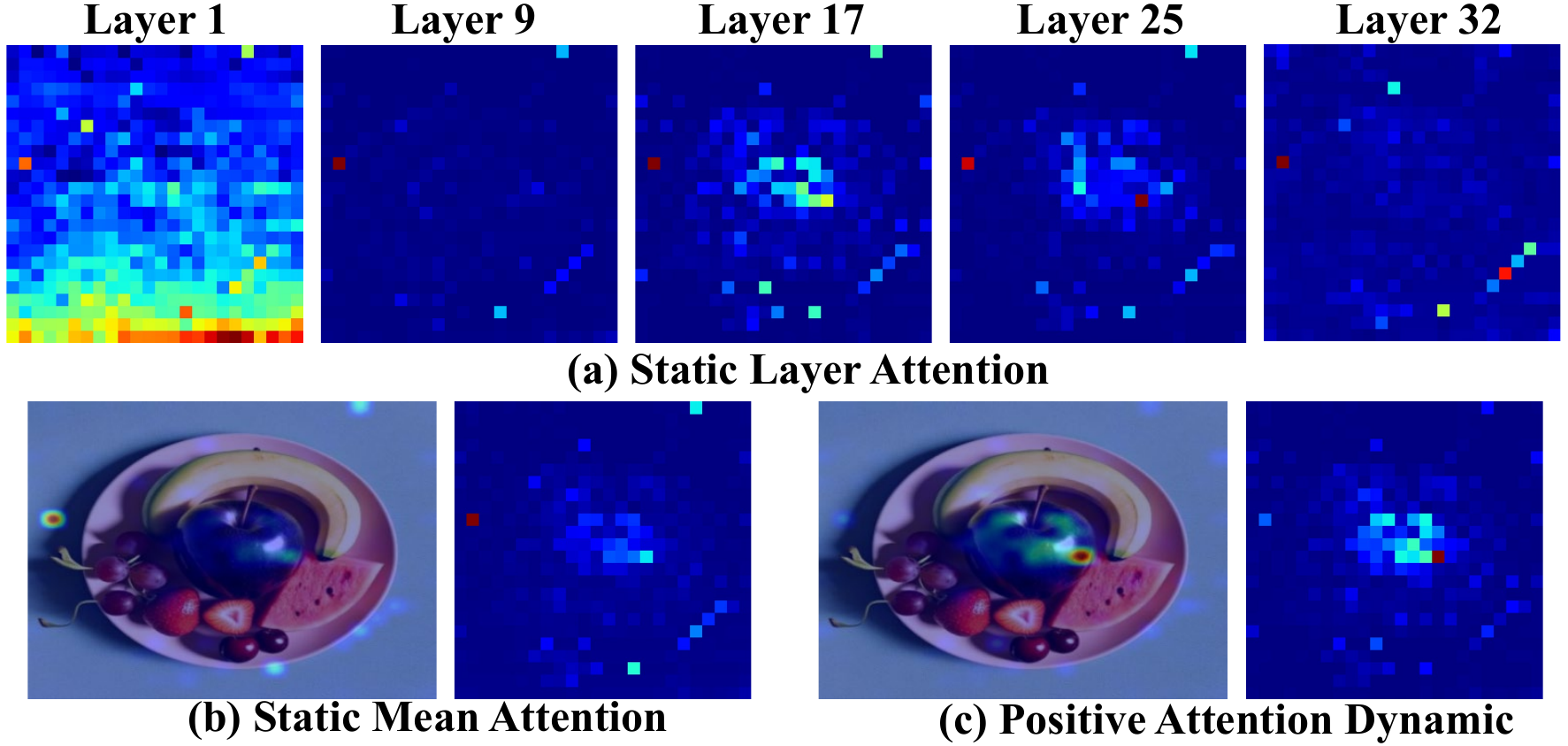}
    \end{minipage}
    \begin{minipage}{0.49\textwidth}
        \centering
    \includegraphics[width=0.99\linewidth]{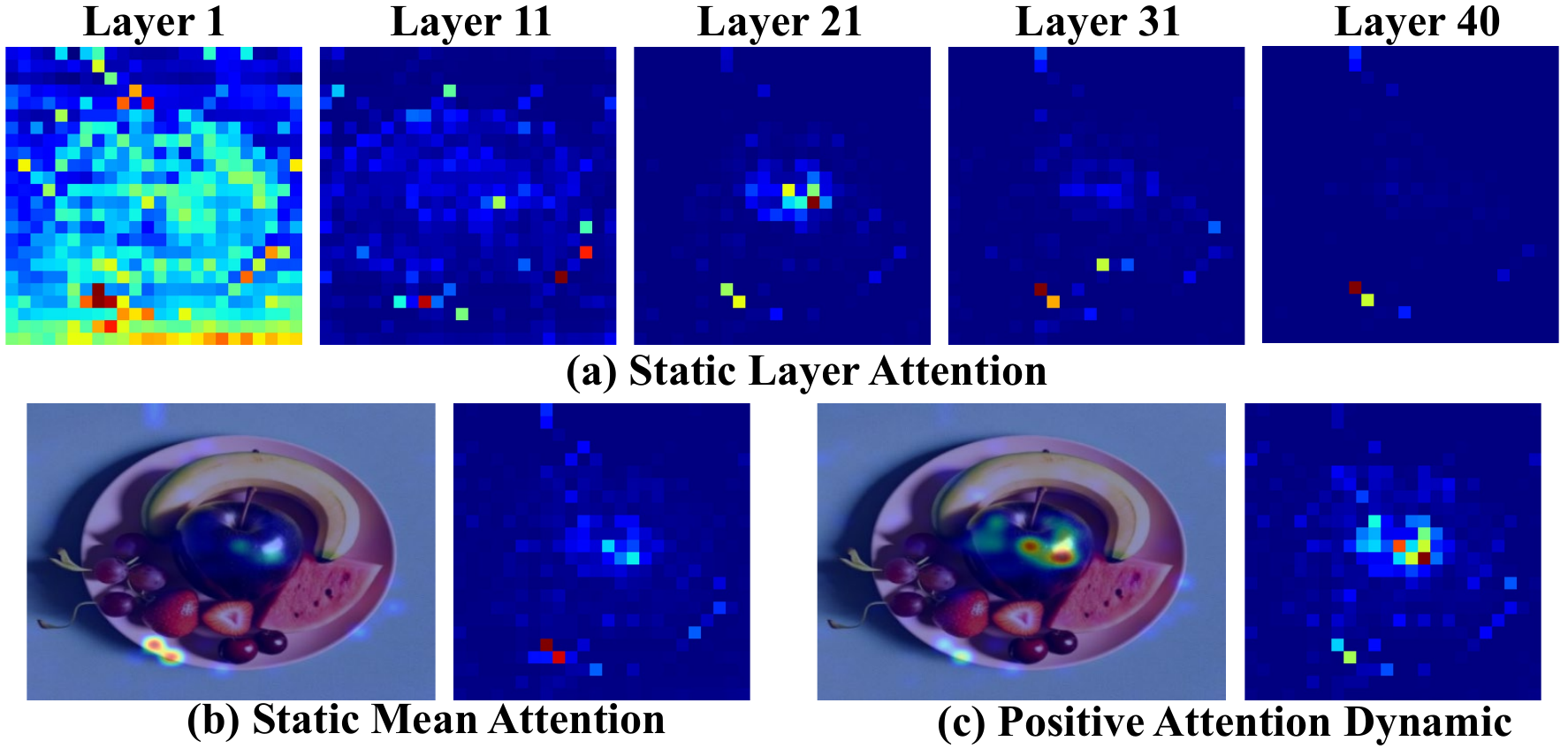}
    \end{minipage}

\caption{Attention analysis of LLaVA-1.5-7B (left) and 13B (right). 
Top: the attention ratio of different token types (System, Vision, Instruction, Output). 
Bottom: heatmap visualizations of attention distributions, including (a) static attention from uniformly sampled layers, (b) layer-averaged static attention, and (c) positive attention dynamics.}
\label{fig:analyse_attn_ratio_heatmap}
\end{figure*}

\section{Related Work}\label{sec:related-work}

\noindent\textbf{LVLMs.}
LVLMs~\cite{instructblip,llava,qwenvl,hu2024minicpm,llavanext,chen2024internvl,yang2025qwen3,llama,vicuna,deepseek,deepseekr1} achieve strong performance across a wide range of vision-language tasks~\cite{jiang2024senna_driving,hu2024omnimedvqa_medical,minigpt4,tmr_guangtao_1,chenghao_2,zhang_peirong2025aesthetics,lin2024videollava}.
Despite this progress, hallucination remains a fundamental challenge for LVLMs~\cite{lee2018hallucinations,gunjal2024detecting,liu2024survey_lvlm,woo2025miss_forest_tree_attn_vision_calibration}.

\noindent\textbf{Hallucinations in LVLMs.}
Existing mitigation approaches can be broadly categorized into two classes.
\textbf{Training-based methods} aim to reduce hallucinations by strengthening modality alignment and robustness, typically through improved data curation, alignment objectives, retrieval-augmented generation, or reinforcement learning ~\cite{liumitigating_data_finetune,yu2024rlhfv_finetune,yu2024rlaifv_finetune,chenperturbollava_finetune,ouali2024clip_clipdpo_finetune,chenperturbollava_finetune}. These approaches require substantial computational resources and retraining costs, limiting their practicality and flexibility.
\textbf{Training-free methods} intervene directly in the decoding process by manipulating logits, attention, or hidden states, and can be grouped into three categories ~\cite{adhh,zhang2024seeing_eah,second_lack_visual,code,huo2024self_sid_lack_visual,zhouanalyzing_lure,lyu2026towards_guangtao_lmm_sae,lyu2025revealing_guangtao_lmm_vdc,li2025visual_hallu_acl,zhang2025cchall_hallu_acl,huangyw2025dynamic_hallu_acl,huang2025alleviating_hallu_acl,tang2025seeing_hallu_attn,hallu_attention_lens,register_dont_need,register_need,Seeing_but_Not_Believing,Seeing_but_Not_Believing_4,Seeing_but_Not_Believing_1,wang2024mllm_hallu_mllm_see,xing2024mitigating_hallu_attn,jianginterpreting_hallu_logits_lens}.
(1) \textit{Contrastive decoding} contrasts outputs under different visual conditions, such as PAI~\cite{pai_lack_visual}, IBD~\cite{zhu2024ibd}, and VCD~\cite{vcd_lack_visual}. These methods require multiple forward passes and may introduce bias from the contrastive perturbed signal.
(2) \textit{Auxiliary expert models} leverage external models to provide auxiliary cues (e.g., HALC~\cite{halc_lack_visual}, AGLA~\cite{an2025mitigating_agla__lack_visual}, Woodpecker~\cite{woodpecker}), which over-rely on external models and may not be aligned with the target LVLM.
(3) \textit{Static internal signal} methods select and amplify vision-related heads, layers, or tokens based on heuristic scores or internal signals (e.g., VAR~\cite{kang2025see_attention_sinks_var}, VAF~\cite{yin2025clearsight_vaf}, OPERA~\cite{opera_lack_visual}, MemVR~\cite{dola_looking_twice_memvr_ffn_iternal}). These methods are vulnerable to the \emph{attention sink or massive activations}~\cite{sun2024massive_activations_attention_sinks,kang2025see_attention_sinks_var}, where semantically irrelevant but dominant tokens are repeatedly amplified. Concurrent work GIFT~\cite{qi2025capturing_gaze_gift} also leverages attention variations, but it relies on part-of-speech analysis to identify key words, which may fail for generic prompts without explicit key objects (e.g., “Please describe this image in detail”), and selects specific visual heads for intervention. In contrast, PADE extracts semantic core regions directly from the LVLM's internal attention dynamics, uses MAD to adaptively control intervention strength, and employs STC to preserve both complex user instruction and historical long-term outputs.

\begin{figure*}[t]
\centering
\includegraphics[width=0.99\linewidth]{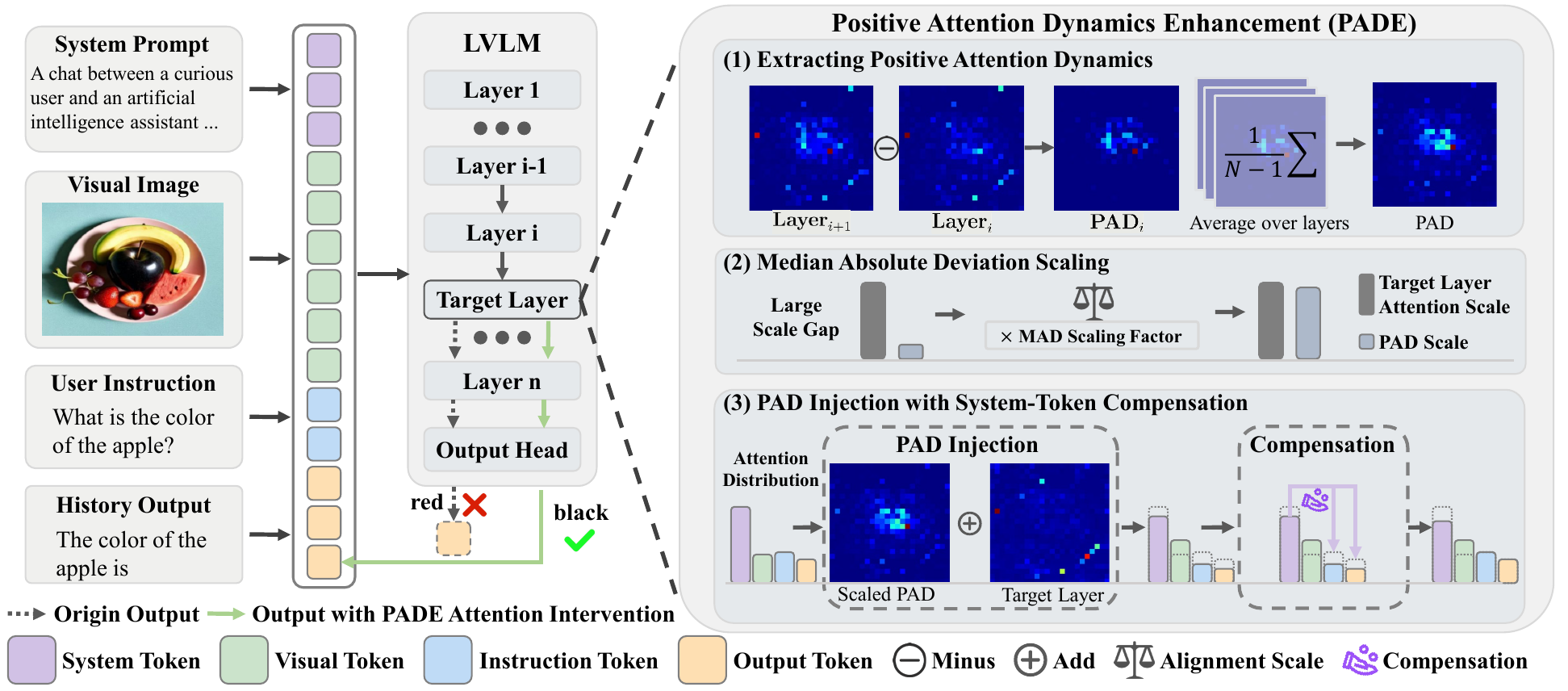}
\caption{Overview of our PADE. 
PADE identifies semantically core visual regions via Positive Attention Dynamics (PAD) and selectively enhances them in the target layer, with Median Absolute Deviation Scaling for adaptively controlling the intervention strength and System-Token Compensation to preserve attention for understanding complex instructions and ensuring consistent long-term generation.
}
\label{fig:framework}
\end{figure*}

\section{Analysis: Internal Attention Dynamics}
\label{sec:analysis}

In this section, we analyze the internal attention behaviors of LVLMs to understand how attention is distributed and evolves across layers during multimodal understanding and reasoning.   
We systematically study  different token types attention distributions and visual attention evolution in LLaVA-1.5 models at two scales (7B and 13B) , as shown in~\Cref{fig:analyse_attn_ratio_heatmap}, with additional qualitative examples provided in~\Cref{sec:app_more_examples} under diverse visual contexts and prompt variations.

\subsection{Attention Sink Dominance in LVLMs}
As shown in ~\Cref{fig:analyse_attn_ratio_heatmap}, 
both LLaVA-1.5-7B and 13B exhibit a highly imbalanced attention distribution, 
where a small set of tokens consistently absorb a disproportionate amount of attention mass. 
These tokens, commonly referred to as \emph{attention sinks}~\cite{xiao2023efficient_attention_sinks_ori,kang2025see_attention_sinks_var,sun2024massive_activations_attention_sinks}, persist across layers 
despite limited semantic relevance to the visual content.

\subsection{Static Metrics Are Vulnerable to Sinks}

Attention sinks manifest as isolated extreme activations with large magnitudes, 
absolute attention maps become heavily skewed. 
Semantically irrelevant sink tokens are often assigned high attention scores, 
while genuinely informative visual regions are comparatively suppressed.
This distortion fundamentally limits prior attention intervention methods~\cite{yin2025clearsight_vaf,only_lack_visual} 
that rely on internal static signals.
Since sink tokens frequently rank among the highest-attended elements, 
they are repeatedly selected and further amplified, 
biasing intervention toward spurious structures and degrading visual grounding reliability.

\subsection{Attention Dynamics Reveal Core Regions}

As illustrated in \Cref{fig:analyse_attn_ratio_heatmap}, we observe distinguishable attention dynamics among semantically core regions, attention sinks, and irrelevant background areas.
Semantically meaningful regions exhibit multiple instances of pronounced positive attention changes across layers, even though their attention may occasionally decrease. 
In contrast, most irrelevant regions maintain consistently low attention with only minor fluctuations. 
Attention sink tokens can display large or sporadic spikes, but these changes are irregular and do not align consistently with semantic understanding and reasoning.

Motivated by these observations, we quantify and highlight semantically core visual regions using
\emph{Positive Attention Dynamics (PAD)}, 
computed as the \emph{positive inter-layer attention deltas} between consecutive layers. 
Semantically core regions often exhibit repeated substantial increases in attention, 
which typically correspond to decreases in other regions. 
By retaining only positive deltas and differential formulation, PAD naturally suppresses attention in less relevant areas, 
and inherently mitigates the influence of most attention sinks. 
As a result, PAD provides a more reliable signal for identifying semantically core visual regions than static attention maps, 
by leveraging the model's internal attention evolution, without requiring external models or auxiliary cues (\Cref{fig:motivation_pad_vs_static,fig:analyse_attn_ratio_heatmap}).

\section{Method: PADE}

Building on our analysis, we propose Positive Attention Dynamics Enhancement (PADE), a training-free attention intervention that selectively reinforces semantically core visual regions to improve visual grounding and mitigate hallucinations in LVLMs. PADE identifies core regions by leveraging the model’s internal \emph{positive attention dynamics} and enhances them in the target layer, without relying on external models or multiple forward passes. As illustrated in \Cref{fig:framework}, PADE consists of three key steps:
(1) extracting \emph{Positive Attention Dynamics (PAD)} to identify semantically core visual regions,
(2) applying per-head \emph{Median Absolute Deviation (MAD)} scaling to adaptively control intervention strength, and
(3) introducing \emph{System-Token Compensation (STC)} to preserve attention to complex instructions and 
long-term generation.

\subsection{Positive Attention Dynamics}

To enable effective attention intervention, we first identify semantically core visual regions that should be selectively reinforced.
Following our analysis in ~\Cref{sec:analysis}, such core visual regions tend to receive {positive} attention increases as the model refines its internal understanding, whereas irrelevant regions rarely exhibit small changes and attention sinks show irregular spikes.
Motivated by this observation, we extract {Positive Attention Dynamics (PAD)}, which aggregates positive inter-layer attention deltas to reveal semantically core visual regions.
By retaining only positive deltas, PAD emphasizes regions whose importance increases during reasoning, while naturally suppressing noisy fluctuations and attention sinks.

Let $\mathbf{A}_l$ denote the visual attention map at layer $l$, averaged over all attention heads.
The positive inter-layer attention delta is defined as
\begin{equation}
\Delta^+ \mathbf{A}_l = \max\bigl(0,\, \mathbf{A}_l - \mathbf{A}_{l-1}\bigr), \quad l = 2, \dots, L .
\end{equation}
We aggregate these deltas across layers to obtain the PAD:
\begin{equation}
\mathbf{P} = \frac{1}{L-1} \sum_{l=2}^{L} \Delta^+ \mathbf{A}_l .
\end{equation}

\subsection{Per-Head MAD Scaling}

The PAD is injected into attention {logits} rather than post-softmax attention, preserving the inherent properties of attention and avoiding an additional softmax operation. 
However, attention logits often contain extreme outliers induced by sink tokens and operate at a much larger scale than PAD, with significant variation across samples. Without proper scaling, the same intervention coefficient $\lambda$ can produce inconsistent perturbation magnitudes, leading to poorly calibrated interventions.

We adaptively control the intervention strength by scaling each attention head with the median absolute deviation (MAD), which uses the median instead of the mean to provide robust per-head, per-sample calibration and reduce the influence of extreme attention sink values.

Let $\mathbf{Z}^{v}_{l,h}$ denote the visual attention logits of head $h$ at layer $l$. The MAD is computed as
\begin{align}
\mathrm{MAD}(\mathbf{Z}^{v}_{l,h}) 
&= \mathrm{median}\!\left(
      \left| \mathbf{Z}^{v}_{l,h} - \mathrm{median}(\mathbf{Z}^{v}_{l,h}) \right|
   \right),
\end{align}
and the PAD is scaled accordingly:
\begin{align}
\hat{\mathbf{P}}_{l,h} 
&= \mathrm{MAD}(\mathbf{Z}^{v}_{l,h}) \cdot \tilde{\mathbf{P}} .
\end{align}

\subsection{System-Token Compensation (STC)}

After scaling PAD to match the magnitude of attention logits, we inject it into the visual attention logits of the target layer:
\begin{equation}
\hat{\mathbf{Z}}^{v}_{l,h} \leftarrow \mathbf{Z}^{v}_{l,h} + \lambda \cdot \hat{\mathbf{P}}_{l,h},
\end{equation}
where $\lambda$ controls the intervention strength.

Directly increasing visual attention will reduce attention to user instructions or previously generated outputs, potentially impairing instruction following and output coherence for complex or long-form tasks. Formally, attention is computed over different token groups as
\begin{equation}
\mathbf{A} = \mathrm{softmax}\bigl([\mathbf{Z}^{s}, \mathbf{Z}^{v}, \mathbf{Z}^{i}, \mathbf{Z}^{o}]\bigr),
\end{equation}
where $\mathbf{Z}^{s}$, $\mathbf{Z}^{v}$, $\mathbf{Z}^{i}$, and $\mathbf{Z}^{o}$ denote the logits of system tokens, user instruction tokens, visual tokens, and output tokens, respectively. 

As shown in \Cref{fig:analyse_attn_ratio_heatmap}, system tokens consistently receive a large proportion of attention and remain largely unrelated to the semantic content of user instructions or visual inputs. Based on this observation, we introduce {System-Token Compensation (STC)}, which leverages the high-attention but semantically irrelevant system tokens to compensate for the increased visual attention, rather than affecting instruction or history tokens. Specifically, system-token logits are adjusted as
\begin{equation}
\check{\mathbf{Z}}^{s} \leftarrow \mathbf{Z}^{s} - \mathrm{mean}\!\left(\lambda \cdot \hat{\mathbf{P}}_{l,h}\right),
\end{equation}
allowing selective enhancement of semantically core visual regions while preserving attention to user instructions and previously generated outputs. The final attention weights are computed as 
\begin{equation}
\hat{\mathbf{A}} = \mathrm{softmax}\bigl([\check{\mathbf{Z}}^{s}, \hat{\mathbf{Z}}^{v}, \mathbf{Z}^{i}, \mathbf{Z}^{o}]\bigr).
\end{equation}

\begin{table*}[t]
    \caption{Results on the POPE (Accuracy and F1). $\uparrow$ indicates that higher is better. Best results are \textbf{bolded}.}
    \label{tab:results_pope}
    \centering
    \resizebox{\textwidth}{!}{
    \begin{tabular}{cl cc cc cc cc cc}
    \toprule
      \multirow{2}{*}{Setup} & \multirow{2}{*}{Method} 
     & \multicolumn{2}{c}{LLaVA-1.5} 
     & \multicolumn{2}{c}{InstructBLIP} 
     & \multicolumn{2}{c}{Qwen-VL}
     & \multicolumn{2}{c}{LLaVA-1.5-13B}
     & \multicolumn{2}{c}{LLaVA-Next} \\
    \arrayrulecolor{gray}
    \cmidrule(lr){3-4} \cmidrule(lr){5-6} \cmidrule(lr){7-8} \cmidrule(lr){9-10} \cmidrule(lr){11-12}
     &   & Accuracy $\uparrow$ & F1 $\uparrow$ 
       & Accuracy $\uparrow$ & F1 $\uparrow$ 
       & Accuracy $\uparrow$ & F1 $\uparrow$ 
       & Accuracy $\uparrow$ & F1 $\uparrow$ 
       & Accuracy $\uparrow$ & F1 $\uparrow$ \\
    \midrule

\multirow{5}{*}{Random}
& Vanilla      & 84.63 & 84.99 & 83.33 & 83.57 & 85.17 & 83.00 & 83.27 & 84.27 & 84.23 & 82.14 \\
& VCD          & 84.57 & 85.02 & 84.60 & 84.49 & 84.69 & 82.91 & 83.47 & 84.52 & 83.48 & 81.36 \\
& PAI         & 85.12 & 85.64 & 83.82 & 83.98 & 85.63 & 83.74 & 84.06 & 84.98 & 83.76 & 81.52 \\
& VAF         & 85.64 & 85.82 & 85.12 & 85.28 & 85.93 & 84.08 & 84.32 & 85.28 & 84.84 & 82.64 \\
& VAR         & 86.12 & 86.56 & 85.62 & 85.71 & 86.54 & 84.66 & 84.82 & 85.76 & 85.12 & 83.26 \\
& PADE (ours)         & \textbf{86.96} & \textbf{87.42} & \textbf{86.52} & \textbf{86.82} & \textbf{87.14} & \textbf{85.68} & \textbf{86.08} & \textbf{87.16} & \textbf{86.24} & \textbf{84.74} \\

\midrule

\multirow{5}{*}{Popular}
& Vanilla      & 81.33 & 82.33 & 76.00 & 77.94 & 84.50 & 82.50 & 80.57 & 82.19 & 82.33 & 80.44 \\
& VCD          & 81.57 & 83.02 & 76.68 & 77.82 & 84.37 & 82.46 & 80.96 & 82.47 & 82.46 & 80.68 \\
& PAI         & 81.82 & 83.46 & 76.82 & 78.56 & 84.83 & 82.76 & 80.78 & 82.74 & 82.76 & 80.82 \\
& VAF         & 82.62 & 84.18 & 77.36 & 79.02 & 86.12 & 84.28 & 81.62 & 83.46 & 83.57 & 81.68 \\
& VAR         & 83.16 & 85.52 & 78.12 & 79.96 & 86.89 & 84.92 & 82.74 & 84.62 & 84.62 & 82.74 \\
& PADE (ours)        & \textbf{84.56} & \textbf{86.28} & \textbf{78.74} & \textbf{80.76} & \textbf{87.72} & \textbf{86.12} & \textbf{83.82} & \textbf{85.28} & \textbf{85.56}  & \textbf{84.12} \\

\midrule

\multirow{5}{*}{Adversarial}
& Vanilla      & 75.87 & 78.27 & 74.17 & 76.58 & 82.53 & 80.56 & 75.12 & 78.35 & 79.37 & 77.88 \\
& VCD          & 75.76 & 78.12 & 74.62 & 76.72 & 82.68 & 80.78 & 75.46 & 78.42 & 79.56 & 78.12 \\
& PAI         & 76.12 & 78.53 & 74.82 & 76.96 & 82.83 & 80.82 & 75.72 & 78.64 & 79.68 & 78.14 \\
& VAF         & 76.72 & 79.14 & 75.48 & 77.73 & 83.52 & 81.38 & 76.63 & 79.12 & 80.23 & 78.74 \\
& VAR        & 77.43 & 80.16 & 76.22 & 78.16 & 83.96 & 81.82 & 77.52 & 79.46 & 80.92 & 79.56 \\
& PADE (ours)         & \textbf{78.47} & \textbf{81.12} & \textbf{77.26} & \textbf{79.13} & \textbf{85.12} & \textbf{83.08} & \textbf{78.62} & \textbf{80.62} & \textbf{81.08} & \textbf{80.92} \\

\bottomrule
    \end{tabular}
}

\end{table*}

\begin{table*}[t]
    \caption{{Results on CHAIR (Max Token 128).} $\downarrow$ denotes lower is better.  -- denotes unavailable results.}
    \label{tab:results_chair}
    \centering
    \resizebox{\textwidth}{!}{
    \begin{tabular}{lcc cc cc cc cc}
        \toprule
        \multirow{2}{*}{Method} 
        & \multicolumn{2}{c}{LLaVA-1.5} 
        & \multicolumn{2}{c}{InstructBLIP} 
        & \multicolumn{2}{c}{Qwen-VL}
        & \multicolumn{2}{c}{LLaVA-1.5-13B}
        & \multicolumn{2}{c}{LLaVA-Next} \\
        \cmidrule(lr){2-3} \cmidrule(lr){4-5} \cmidrule(lr){6-7} \cmidrule(lr){8-9} \cmidrule(lr){10-11}
        & CHAIR$_S$ $\downarrow$ & CHAIR$_I$ $\downarrow$ 
        & CHAIR$_S$ $\downarrow$ & CHAIR$_I$ $\downarrow$ 
        & CHAIR$_S$ $\downarrow$ & CHAIR$_I$ $\downarrow$
        & CHAIR$_S$ $\downarrow$ & CHAIR$_I$ $\downarrow$
        & CHAIR$_S$ $\downarrow$ & CHAIR$_I$ $\downarrow$ \\
        \midrule

Vanilla   & 55.1 & 16.4 & 57.4 & 17.6 & 52.1 & 16.7 & 50.4 & 14.7 & 30.2 & 10.9 \\
VCD       & 54.4 & 16.6 & 60.7 & 18.0 & 50.4 & 17.2 & 49.6 & 14.3 & 29.8 & 10.6 \\
M3ID    & 56.6 & 15.8 & 62.3 & 18.2 & 49.8 & 17.4 & — & — & — & — \\
Woodpecker& 57.6 & 16.7 & 60.8 & 17.6 & 51.8 & 16.3 & — & — & — & — \\
HALC      & 51.0 & 14.8 & 53.8 & 15.7 & 49.6 & 15.4 & — & — & — & — \\
ONLY      & 49.8 & 14.3 & 52.2 & 15.5 & 48.0 & 14.3 & 48.9 & 13.6 & 28.6 & 9.7 \\
AGLA      & 52.4 & 14.6 & 54.8 & 16.2 & 49.8 & 15.6 & — & — & — & — \\
OPERA     & 51.6 & 14.2 & 54.2 & 14.8 & 48.6 & 14.6    & — & — & — & — \\

PAI  & 53.1 & 15.1 & 54.2 & 15.6 & 49.1 & 15.6 & 49.4 & 14.1 & 29.4 & 10.4 \\
VAF  & 50.1 & 14.2 & 53.4 & 15.1 & 48.7 & 14.4 & 48.6 & 13.4 & 28.4 & 9.6 \\
VAR      & 49.6 & 14.1 & 52.3 & 14.7 & 48.4 & 13.9 & 48.2 & 13.2 & 28.2 & 9.2 \\
PADE (ours)    & \textbf{48.6} & \textbf{13.7} & \textbf{51.8} & \textbf{14.2} & \textbf{47.8} & \textbf{13.4} & \textbf{47.3} & \textbf{12.6} & \textbf{27.8} & \textbf{8.9} \\
        \bottomrule
    \end{tabular}
    }
\end{table*}

\begin{table*}[t]
\centering
\caption{Results on multiple general vision language benchmarks. $\uparrow$ indicates that higher is better.}
\label{tab:results_general_benchmarks}
        \resizebox{1\linewidth}{!}{
\begin{tabular}{l c c c c c c}
\toprule
\multirow{2}{*}{Methods} 
& VizWiz
& \multicolumn{3}{c}{MME} 
& LLaVA-Wild 
& MM-Vet \\
\cmidrule(lr){2-2} 
\cmidrule(lr){3-5} 
\cmidrule(lr){6-6} 
\cmidrule(lr){7-7} 
& Accuracy $\uparrow$ 
& Perception $\uparrow$ 
& Cognition $\uparrow$ 
& Overall $\uparrow$ 
& Average $\uparrow$ 
& Total $\uparrow$ \\
\midrule

Vanilla   & 50.00 & 1508.97 & 355.71 & 1864.68   & 64.80 & 31.1 \\
VCD      & 44.90 & 1515.01 & 357.86 & 1872.87   & 63.21 & 30.2 \\
ICD  & 37.62 & 1306.91 & 287.86 & 1594.77   & 56.90 & 25.9 \\
OPERA   & 50.76 & 1473.62 & 310.71 & 1784.34   & 64.31 & 32.0 \\
INTER  & 48.77 & 1502.35 & 336.18 & 1838.53   & 61.70 & 30.9 \\
V-ITI     & 51.72 & 1518.32 & 369.03 & 1887.35   & 65.44 & 31.7 \\
PADE (ours)    & \textbf{52.08} & \textbf{1520.68} & \textbf{371.44} & \textbf{1892.12}   & \textbf{65.92} & \textbf{32.4} \\

\bottomrule
\end{tabular}
}
\end{table*}

\begin{table}[t]
\centering
\caption{Results on the HallusionBench.}
\label{tab:results_HallusionBench} 

\resizebox{1.0 \linewidth}{!}{
\begin{tabular}{l|ccccc}
\toprule
 Methods & fACC $\uparrow$ & qACC $\uparrow$ & ${easy}$A $\uparrow$ & ${hard}$A $\uparrow$ \\ 
\midrule
Vanilla   & 17.9 & 8.13 & 36.0 & 36.7  \\
VCD       & 13.9 & 11.4 & 33.0 & 34.7  \\
ICD       & 13.9 & 8.35 & 36.9 & 33.5  \\
OPERA     & 16.2 & 5.49 & 37.6 & 35.4  \\
INTER     & 15.8 & 8.21 & 36.9 & 34.5  \\
V-ITI     & 17.9 & 10.27 & 36.6 & 37.0  \\
PADE (ours)     & \textbf{18.1} & \textbf{11.56} & \textbf{37.9} & \textbf{37.4}  \\
\bottomrule
\end{tabular}}
\end{table}

\begin{table}[t]
\caption{Results on AMBER Generative Subset.}
\label{tab:results_amber}
\centering
\resizebox{1\linewidth}{!}{
\begin{tabular}{lcccc}
\toprule
\textbf{Method} & \textbf{CHAIR ($\downarrow$)} & \textbf{Cover ($\uparrow$)} & \textbf{Hall ($\downarrow$)} & \textbf{Cog ($\downarrow$)} \\
\midrule
Vanilla & 7.8 & 51.0 & 36.4 & 4.2 \\
VCD & 7.5 & 50.8 & 36.2 & 4.1 \\
OPERA & 7.3 & 49.6 & 32.0 & 3.5 \\
DoLA & 7.6 & 51.6 & 36.0 & 4.0 \\
PAI & 7.4 & 49.9 & 33.2 & 3.7 \\
PADE (ours)  & \textbf{7.1} & \textbf{51.8} & \textbf{31.4} & \textbf{3.4} \\

\bottomrule
\end{tabular}
}
\end{table}

\begin{table}[t]
\caption{Ablation results on the proposed components.}
\label{tab:abla_pade}
\footnotesize
\centering
\resizebox{1\linewidth}{!}{
\begin{tabular}{l lcc}
\toprule
\textbf{Model} & \textbf{Components} & \textbf{CHAIR$_S$} $\downarrow$ & \textbf{CHAIR$_I$} $\downarrow$ \\
\midrule
\multirow{3}{*}{LLaVA-1.5-7B}
 & PADE              & \textbf{48.6} & \textbf{13.7} \\
 & w/o MAD           & 54.9 & 16.3 \\
 & w/o STC           & 49.2 & 14.0 \\ 
\midrule
\multirow{3}{*}{LLaVA-1.5-13B}
 & PADE              & \textbf{47.9} & \textbf{12.8} \\
 & w/o MAD           & 50.3 & 14.5 \\
 & w/o STC           & 48.6 & 13.3 \\
\midrule
\multirow{3}{*}{LLaVA-1.5-NeXT}
 & PADE              & \textbf{27.8} & \textbf{8.9} \\
 & w/o MAD           & 30.1 & 10.7 \\
 & w/o STC           & 28.3 & 9.2 \\
\midrule
\multirow{3}{*}{Qwen-VL}
 & PADE              & \textbf{47.3} & \textbf{12.6} \\
 & w/o MAD           & 51.8 & 16.5 \\
 & w/o STC           & 48.1 & 13.7 \\
\midrule
\multirow{3}{*}{InstructBLIP}
 & PADE              & \textbf{51.8} & \textbf{14.2} \\
 & w/o MAD           & 57.2 & 17.3 \\
 & w/o STC           & 52.2 & 14.5 \\
\bottomrule
\end{tabular}
}
\end{table}

\section{Experiments}
\label{sec:experiment}

\noindent \textbf{Benchmarks.}  
To comprehensively evaluate PADE, we conduct extensive experiments on two categories of benchmarks: {hallucination-focused} and {general-purpose} multimodal benchmarks. The hallucination-focused benchmarks include POPE~\cite{pope} (binary hallucination classification), CHAIR~\cite{chair} (object hallucination in open-ended captioning), HallusionBench~\cite{hallusionbench} (fine-grained visual consistency), and AMBER~\cite{amber} (visually grounded reasoning and generation). The general-purpose benchmarks include VizWiz~\cite{gurari2018vizwiz}, MME~\cite{mme}, LLaVA-Wild~\cite{llava}, and MM-Vet~\cite{yu2023mmvet}, covering diverse tasks in visual understanding, reasoning, and real-world multimodal scenarios.


\noindent \textbf{Evaluated LVLMs.}
We evaluate our PADE on several representative open-source LVLMs, including {LLaVA-1.5}~\cite{llava}, {InstructBLIP}~\cite{instructblip}, and {Qwen-VL}~\cite{qwenvl}.
We additionally consider a larger scale variant, {LLaVA-1.5-13B}, as well as the stronger and newer {LLaVA-NeXT~\cite{llavanext}}.
Together, these models cover diverse backbone architectures and model scales, providing a comprehensive evaluation setting.
Following prior works~\cite{vcd_lack_visual,only_lack_visual}, we apply sampling-based decoding in default. Unless otherwise specified, {LLaVA-1.5} is used as the default model.

\noindent \textbf{Baselines.}
We compare PADE with various training-free hallucination mitigation methods:
(1) \emph{contrastive decoding} methods (VCD~\cite{vcd_lack_visual}, PAI~\cite{pai_lack_visual}, M3ID~\cite{m3id_lack_visual}, ICD~\cite{wang_icd}, DoLA~\cite{dola}); 
(2) \emph{auxiliary expert model} methods (HALC~\cite{halc_lack_visual}, AGLA~\cite{an2025mitigating_agla__lack_visual}, Woodpecker~\cite{woodpecker}, V-ITI~\cite{sun2025viti}); and 
(3) \emph{static internal signal} methods (OPERA~\cite{opera_lack_visual}, VAF~\cite{yin2025clearsight_vaf}, INTER~\cite{dong2025inter}, ONLY~\cite{only_lack_visual}, VAR~\cite{kang2025see_attention_sinks_var}).

\noindent \textbf{Implementation Details.}  
All experiments are conducted on a single NVIDIA RTX A6000 GPU (48GB). 
Unless otherwise specified, PADE is applied to the final layer with an intervention strength of $\lambda=0.1$. 
PADE introduces negligible computational and memory overhead. 
The method maintains only a single additional attention map and relies on lightweight operations, including inter-layer differencing and MAD scaling.
It requires neither auxiliary models nor multiple forward passes, achieving inference speed comparable to vanilla decoding while improving visual grounding and mitigating hallucinations.

\subsection{Main Experimental Results}

\noindent \textbf{Results on Hallucination Benchmarks.}  
As shown in \Cref{tab:results_pope,tab:results_chair,tab:results_amber,tab:results_HallusionBench}, PADE consistently achieves superior performance across different model architectures and scales on diverse comprehension and generation hallucination benchmarks, demonstrating its effectiveness in mitigating hallucinations.  
Compared with contrastive decoding methods (PAI, VCD, IBD, ICD), which introduce perturbed visual inputs that may disrupt semantic alignment, and auxiliary expert approaches (AGLA, HALC, Woodpecker), which rely on external models or conditions not necessarily aligned with the target LVLM, PADE directly leverages the model’s {internal positive attention dynamics} to identify and reinforce semantically core visual regions.  
In contrast to static internal signal methods (VAF, OPERA, VAR), which are either sensitive to attention sinks or merely reallocate attention, PADE simultaneously enhances the global visual attention ratio while emphasizing semantically core regions, leading to more effective hallucination mitigation.

\noindent \textbf{Results on General Benchmarks.}  
As shown in \Cref{tab:results_general_benchmarks}, many existing hallucination mitigation methods rely on contrastive decoding or auxiliary expert models, which use perturbed images, instructions, or external experts to guide outputs. While effective for hallucination reduction, these approaches often compromise general multimodal understanding and reasoning capabilities, as the interventions are not naturally aligned with the LVLM’s internal reasoning. In contrast, PADE leverages the LVLM’s own internal attention dynamics, selectively reinforcing semantically core visual regions. This allows it to reduce hallucinations while preserving the model’s inherent broad visual reasoning and multimodal understanding, demonstrating a more reliable attention intervention.

\subsection{Ablation Studies}

\noindent \textbf{Effect of Proposed Components.}  
We evaluate the contribution of each component in PADE  on the CHAIR benchmark across multiple LVLMs. 
Specifically, we consider two variants: (i) \emph{w/o MAD} and (ii) \emph{w/o STC}. 
As shown in~\Cref{tab:abla_pade}, removing either component consistently degrades performance, while the full PADE achieves the best results across all models.
Notably, removing MAD leads to a substantial degradation in performance, as the raw PAD signal is several orders of magnitude smaller than the original attention logits. Without proper scaling, the intervention becomes largely ineffective and the model behavior closely resembles the vanilla baseline, underscoring the necessity of robust and adaptive scale calibration.
These results confirm that MAD and STC play roles in ensuring PADE’s effectiveness and stability.

\noindent \textbf{Intervention Layer.}
We study the impact of the intervention layer using LLaVA-1.5 7b and 13b on the CHAIR benchmark.
As shown in ~\Cref{fig:abla_layer_lambda}, the effectiveness of PADE generally improves when applied to later layers, with the best performance achieved at the final layer.
This trend can be attributed to the evolution of visual attention across layers: the model typically attends to semantically core visual regions in intermediate layers, while attention becomes more diffuse and spreads to less relevant regions in later layers.
Although the overall visual attention ratio is already high at the final layer, it is often dominated by non-core regions.
Injecting PADE at this stage effectively re-emphasizes semantically core regions while suppressing dispersed attention, leading to the most pronounced improvement.
Moreover, the final layer naturally allocates less attention to system tokens and more to visual tokens and historical outputs, making them more receptive to targeted visual enhancement.

\noindent \textbf{Intervention Strength $\lambda$.}
We study the effect of the intervention strength $\lambda$ on LLaVA-1.5 7b and 13b using the CHAIR benchmark.
As shown in ~\Cref{fig:abla_layer_lambda}, PADE  achieves better performance with relatively small values of $\lambda$ (e.g., 0.1 and 0.3), with $\lambda$=0.1 performing best overall.
Moderate intervention strengths effectively enhance semantically core visual regions while preserving the original attention distribution, whereas excessively large $\lambda$ introduces overly strong interventions and perturbations that deviate from the model’s learned attention dynamics and degrade performance.

\begin{figure}[t]
    \begin{minipage}{0.49\textwidth}
        \centering
    \includegraphics[width=0.99\linewidth]{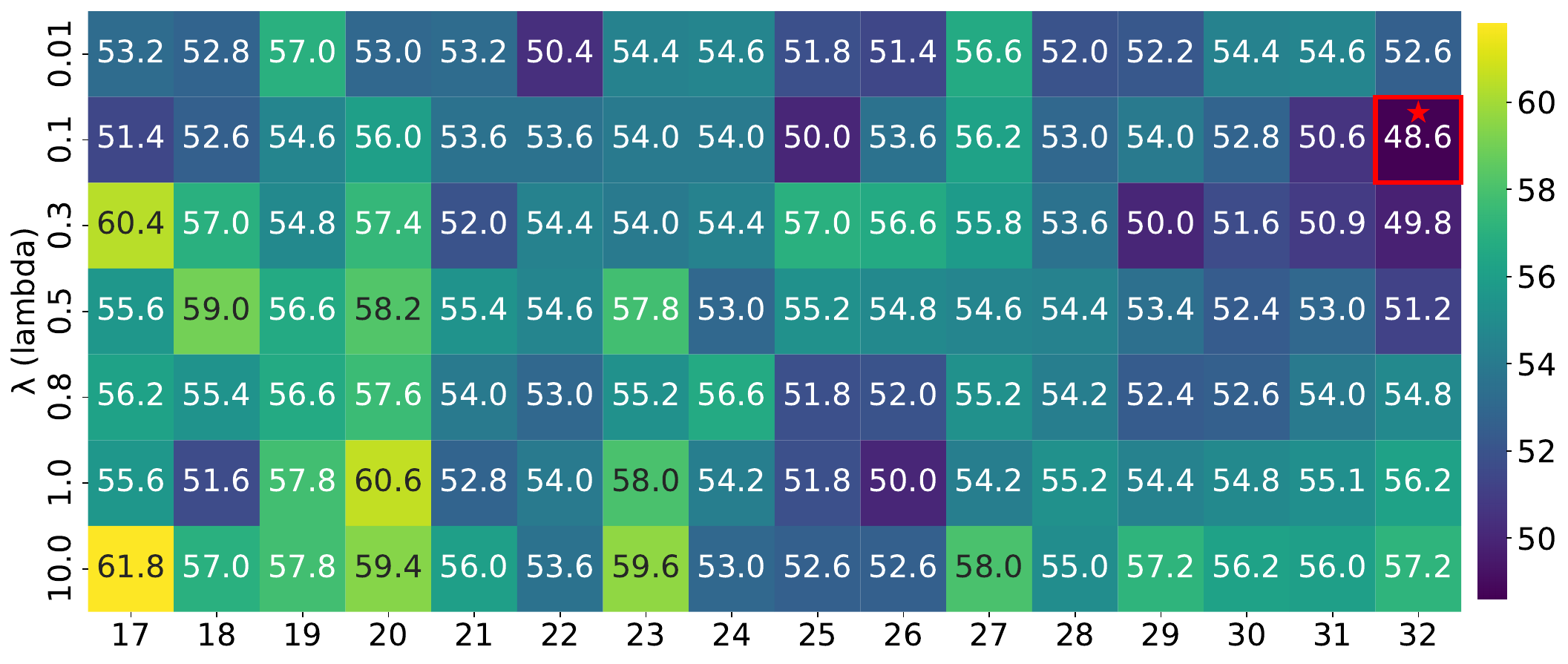}
    \end{minipage}

    \begin{minipage}{0.49\textwidth}
        \centering
    \includegraphics[width=0.99\linewidth]{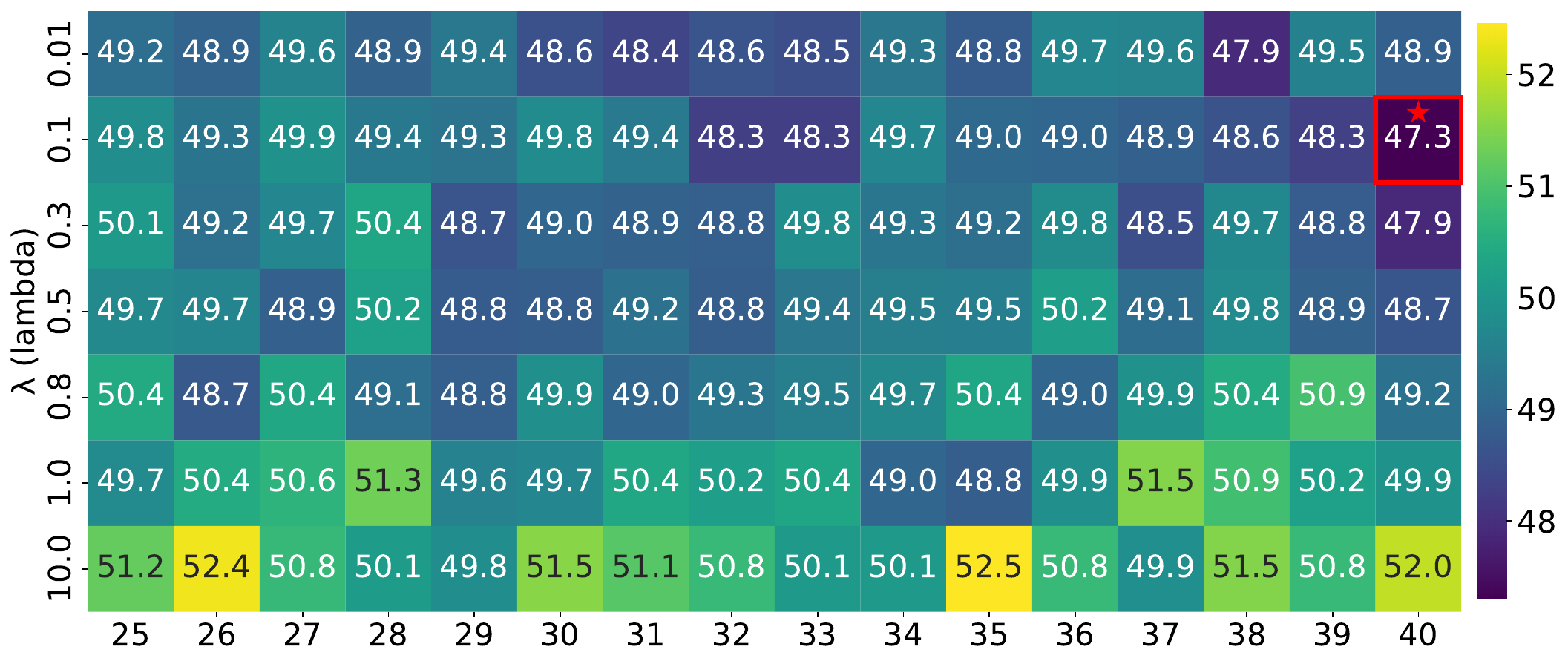}
    \end{minipage}

\caption{Ablation results on the intervention layer and strength $\lambda$ of LLaVA-1.5-7B (top) and 13B (bottom). 
}
    \label{fig:abla_layer_lambda}
\end{figure}

\section{Conclusion}
\label{sec:conclusion}

In this paper, we investigate the role of internal positive attention dynamics in LVLMs and show that they can reveal semantically core visual regions under attention sink distortions. Based on this finding, we propose {Positive Attention Dynamics Enhancement (PADE)}, a lightweight, training-free intervention that constructs a PAD map to identify core visual regions, applies per-head Median Absolute Deviation (MAD) scaling to adaptively adjust the intervention strength, and leverages a System-Token Compensation (STC) module to maintain attention to complex user instructions and support long-term output consistency.  Extensive experiments demonstrate that PADE improves visual grounding and reduces hallucinations, validating the effectiveness of exploiting internal signal dynamics.

\section{Limitations}
\vspace{-4pt}

PADE emphasizes {internal dynamic signals} rather than static magnitude-based criteria, leveraging the evolution of attention across layers to identify semantically core regions. While our analysis demonstrates the effectiveness of attention dynamics for inference-time intervention, the current study focuses exclusively on attention-based mechanisms. Other internal representations, such as the dynamics of hidden states, activation patterns in feed-forward networks, or variations in output logits during decoding, are not explored. These components may also encode complementary signals related to semantic grounding and hallucination behavior. Extending the dynamic analysis framework beyond attention to encompass a broader range of internal model signals remains an important direction for future work.

{
    \bibliography{main}
}
\appendix

\section{Additional Qualitative Examples}
\label{sec:app_more_examples}

We present additional qualitative examples in 
\Cref{fig:app_case_001_run_towards,fig:app_case_002_cup,fig:app_case_003_phone,fig:app_case_003_bag,fig:app_case_different_tokens}
to illustrate how \emph{Positive Attention Dynamics (PAD)} reveal semantically core visual regions under diverse and challenging conditions.
These examples cover direction hallucinations in open-ended long-term generation, object existence hallucinations, atypical or counterintuitive attribute reasoning (e.g., color), as well as small-object and occlusion scenarios across different user prompts.

Across all cases, a common attention pattern emerges in LVLMs: while the model may initially attend to relevant regions in early or intermediate layers, attention progressively diffuses in later layers and becomes dominated by semantically irrelevant sink tokens, which often maintain abnormally high activations across layers. This behavior underscores the unreliability of static attention maps for identifying core visual regions.

Unlike static attention, which is often skewed by attention sinks and fails to highlight small or occluded objects (e.g., \Cref{fig:app_case_002_cup,fig:app_case_003_phone}), PAD captures {positive inter-layer attention deltas} that reliably reflect semantically core regions. By applying PADE at later layers, the model selectively re-emphasizes these core regions (see \Cref{fig:app_case_001_run_towards,fig:app_case_003_bag,fig:app_case_different_tokens}), thereby improving visual grounding and mitigating hallucinations in multimodal understanding.

Together, these examples demonstrate that PAD  highlights semantically core visual regions across varying object scales, scene complexities, and user instructions. In contrast, static attention is often dominated by attention sinks, particularly in later layers, which can obscure meaningful regions.

\begin{figure*}[t]
    \centering
    \includegraphics[width=1\linewidth]{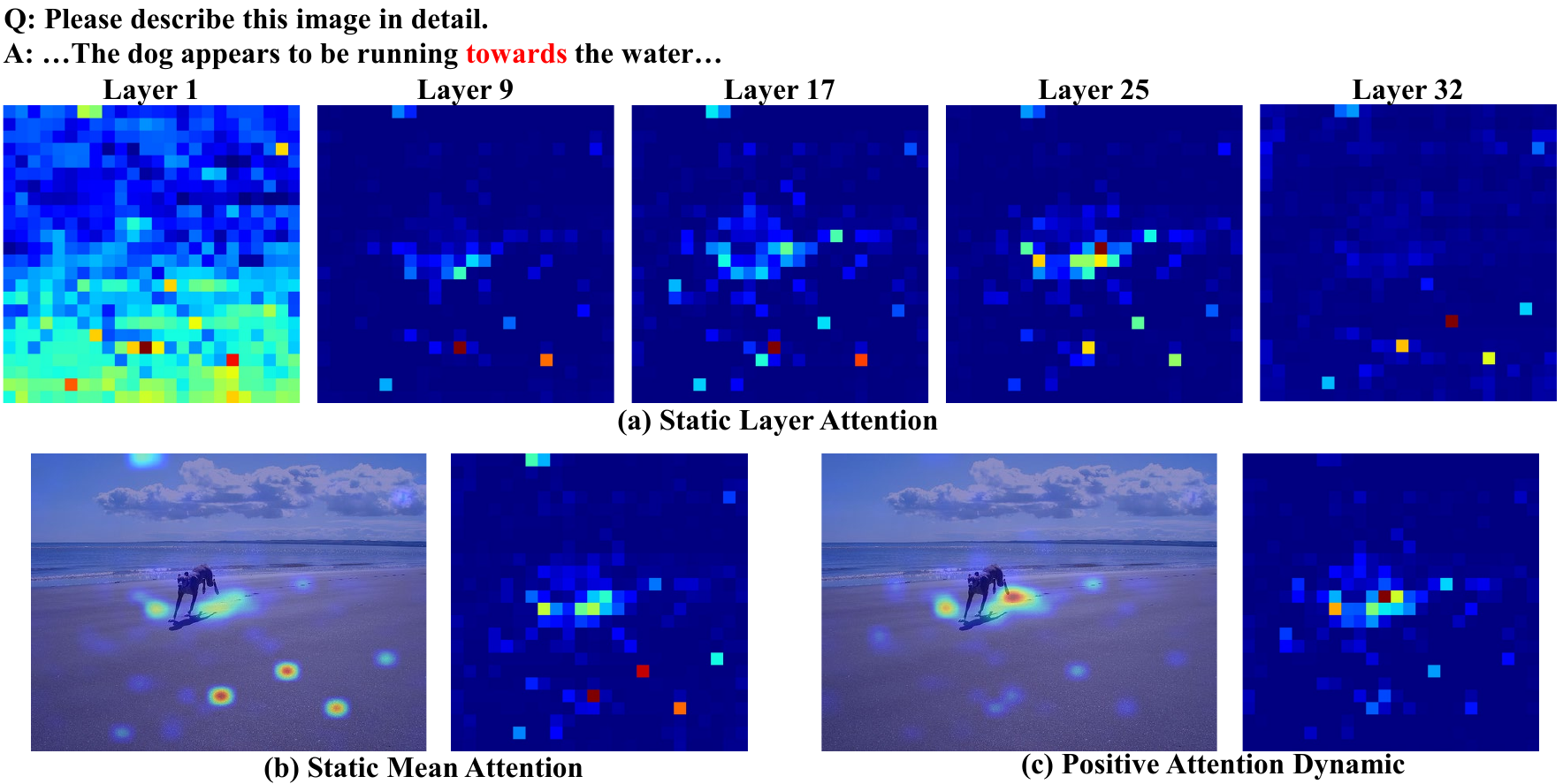}
\caption{
Direction hallucination in open-ended long-term generation.
Although the model attends to the dog throughout generation, it incorrectly describes the running direction as \emph{towards} the water, while the dog is in fact moving away from it.
A substantial portion of attention is absorbed by several semantically irrelevant sink tokens, especially in later layers, diluting the focus on the dog region.
Positive Attention Dynamics (PAD) captures consistent attention changes concentrated around the dog region across layers.
When applied at later layers, PADE selectively reinforces this core region after attention diffusion, correcting the directional hallucination.
}

    \label{fig:app_case_001_run_towards}
\end{figure*}

\begin{figure*}[t]
    \centering
    \includegraphics[width=1\linewidth]{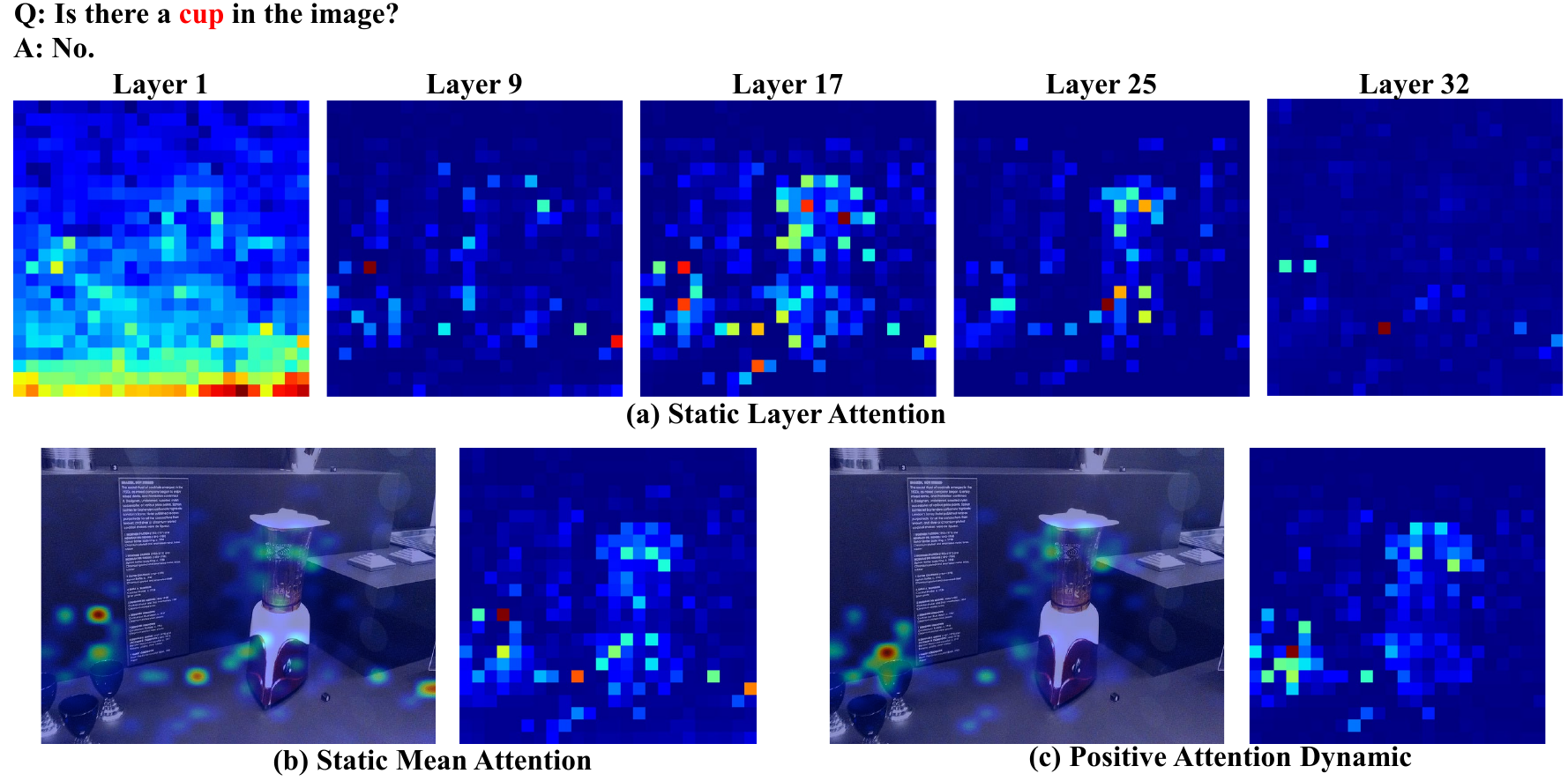}
\caption{
Object existence hallucination in a cluttered scene.
The image contains multiple objects, including several cups and a visually dominant bottle. Due to the presence of multiple salient objects, visual attention becomes fragmented, and smaller target objects such as the cups receive substantially less attention than the large bottle. Although both the cups and the bottle are attended to at intermediate layers, attention gradually diffuses away from semantically relevant regions, and sink tokens increasingly dominate the final-layer attention map, causing the cup regions to become inconspicuous.
PAD captures the transient attention concentration on the cup regions before diffusion occurs, enabling PADE to re-emphasize these small but semantically critical regions in later layers and correctly recognize the existence of cups, thereby mitigating the hallucinated judgment.
}
    \label{fig:app_case_002_cup}
\end{figure*}

\begin{figure*}[t]
    \centering
    \includegraphics[width=1\linewidth]{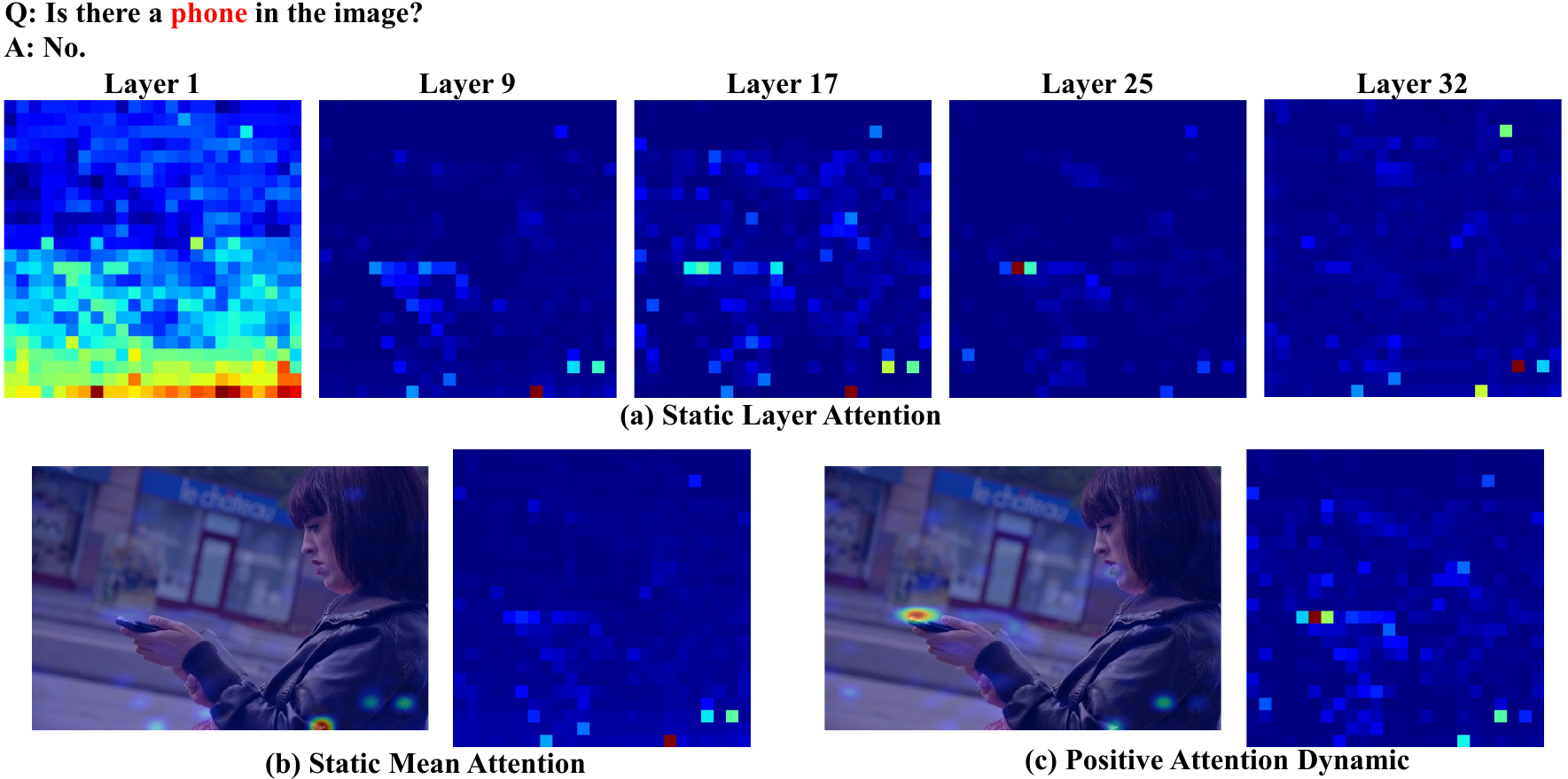}
\caption{
Object existence hallucination for a small target.
The phone occupies a relatively small visual region, resulting in weak and sparse attention signals.
In later layers, attention disperses from the phone region and becomes highly imbalanced, with sink tokens dominating the distribution.
Despite the absence of strong static attention, PAD identifies consistent attention dynamics localized around the phone region.
By reinforcing this region at later layers, PADE restores visual grounding and alleviates the hallucination.
}
    \label{fig:app_case_003_phone}
\end{figure*}

\begin{figure*}[t]
    \centering
    \includegraphics[width=1\linewidth]{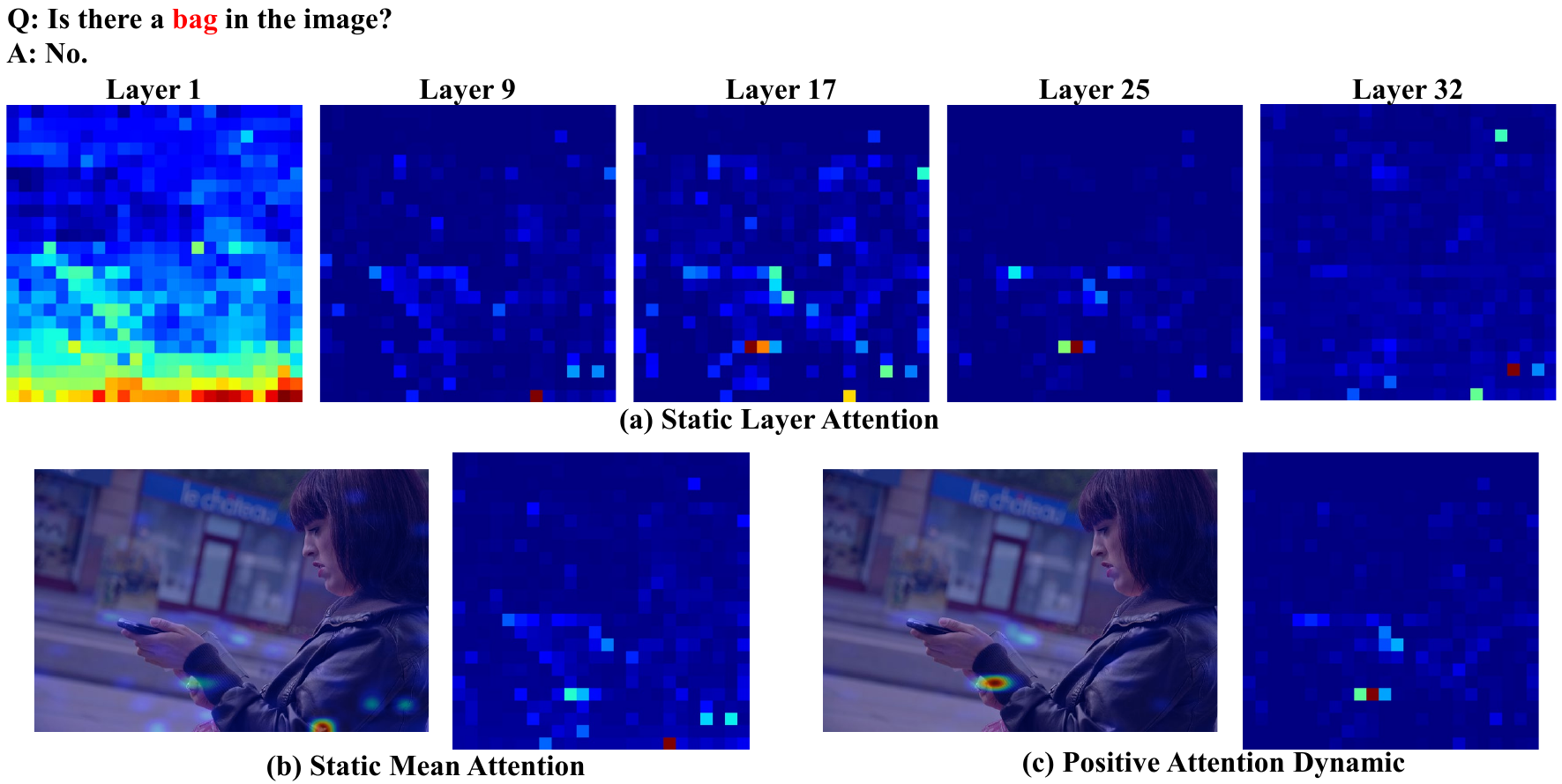}
\caption{
Object existence hallucination under occlusion and prompt variation.
The bag is partially occluded and occupies a small visible area, leading to weak static attention signals.
As attention diffuses in later layers, sink tokens dominate the distribution, resulting in hallucinated responses.
PAD captures coherent attention dynamics around the {hand and bag} regions across layers and enables PADE to selectively enhance these regions in the target layer.
This figure shares the same image as ~\Cref{fig:app_case_003_phone} but uses a different user instruction (bag vs.\ phone).
Both prompts exhibit similar sink-dominated attention patterns and hallucination behaviors, indicating that such failures arise from intrinsic LVLM attention dynamics rather than prompt-specific artifacts.
PADE consistently leverages attention evolution to recover semantically core regions and mitigate hallucinations across prompts.
}
    \label{fig:app_case_003_bag}
\end{figure*}

\begin{figure*}[t]
    \centering
    \includegraphics[width=1\linewidth]{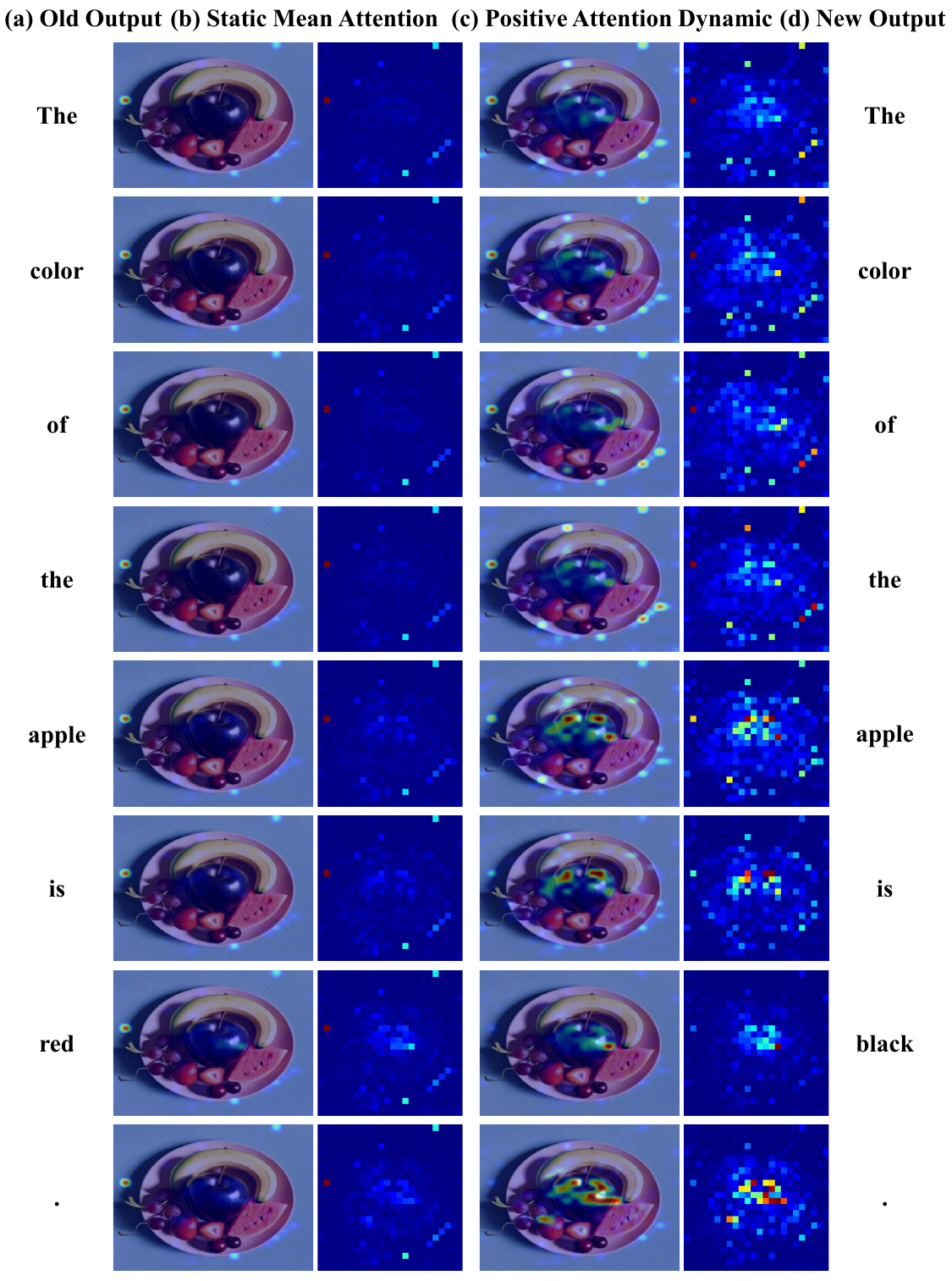}
    \vspace{-14pt}
\caption{
Comparison of static mean attention and positive attention dynamics during the generation of different tokens.
(a) Original model output.
(b) Static mean attention.
(c) Positive Attention Dynamics (PAD).
(d) Output after applying PADE, where semantically core regions are enhanced via PAD.
Across all tokens, PAD more reliably highlights semantically core visual regions than static attention.
In particular, when generating tokens corresponding to key objects (e.g., ``apple'') and attributes (e.g., ``red''), attention changes are most strongly concentrated on the apple region.
These results demonstrate that PAD effectively captures semantically relevant visual areas and validate the effectiveness of PADE in reinforcing core visual regions for reliable multimodal understanding.
}
\label{fig:app_case_different_tokens}
\end{figure*}

\clearpage
\clearpage

\section{Details of Benchmarks}
\label{sub:appendix:benchmark}

To comprehensively evaluate PADE, we conduct experiments on two categories of benchmarks: \emph{hallucination-focused} and \emph{general-purpose} multimodal benchmarks. 
Following prior works~\cite{llava,vcd_lack_visual,sun2025viti}, we adopt standard evaluation protocols for each benchmark to measure both hallucination mitigation and overall multimodal understanding performance. 
Below, we provide a brief overview of each benchmark and its corresponding evaluation focus.

\subsection{Hallucination Benchmarks}
We use the following benchmarks to evaluate the visual hallucination performance:

\begin{itemize}

    \item 
 \textbf{CHAIR}~\cite{chair}:
    CHAIR evaluates object hallucination in open-ended captioning and generation by comparing objects mentioned in model outputs with ground-truth objects present in the image.
    We report both instance-level and sentence-level metrics:
    \begin{align}
    \text{CHAIR}_I &= \frac{|\{\text{hallucinated objects}\}|}{|\{\text{all mentioned objects}\}|}, \\
    \text{CHAIR}_S &= \frac{|\{\text{hallucinated captions}\}|}{|\{\text{generated captions}\}|}.
    \end{align}
    Here, $\text{CHAIR}_I$ measures the proportion of hallucinated objects among all mentioned objects,
    while $\text{CHAIR}_S$ measures the fraction of generated outputs that contain at least one hallucinated object.
    Lower values indicate fewer hallucinations.
    
     \item 
\textbf{POPE}~\cite{pope}:
    POPE (Polling-based Object Probing Evaluation) assesses object existence hallucination via binary (yes/no) questions.
    Queries are evenly split between existent and non-existent objects under random, popular, and adversarial settings.
    We report Accuracy, Precision, Recall, and F1 score, where higher values indicate better object existence discrimination.

    \item 
\textbf{HallusionBench}~\cite{hallusionbench}:
    HallusionBench evaluates fine-grained visual hallucinations in open-ended multimodal responses.
    It reports question-level accuracy ({qACC}) and full-response accuracy ({fACC}), measuring whether answers are fully supported by visual evidence.
    In addition, {easyA} and {hardA} separately evaluate performance on visually simple and challenging cases, respectively.
    Higher scores indicate better visual grounding and hallucination robustness.

    \item 
\textbf{AMBER}~\cite{amber}:
    AMBER focuses on open-ended hallucinations in multimodal reasoning and generation.
    It reports multiple metrics, including {CHAIR}-style object hallucination rates, {Coverage} (Cover), and overall {Hallucination} rate.
    It further distinguishes {Cognitive Hallucination (Cog.)}, which measures hallucinations arising from incorrect reasoning rather than missing visual evidence.
    Lower hallucination-related scores indicate better performance.

\end{itemize}

\subsection{General Benchmarks}
To examine whether mitigating hallucinations comes at the cost of general multimodal understanding, we further evaluate PADE on a suite of general-purpose benchmarks, following prior work~\cite{llava,sun2025viti}:

\begin{itemize}

    \item 
\textbf{VizWiz}~\cite{gurari2018vizwiz}:
    VizWiz is a visual question answering benchmark consisting of images captured by blind or low-vision users.
    The dataset contains diverse real-world challenges such as poor lighting, blur, and occlusion.

     \item 
\textbf{MME}~\cite{mme}:
    MME is a comprehensive evaluation benchmark for multimodal large language models, covering both perception and reasoning abilities.
    It includes tasks such as object existence, counting, spatial relations, color recognition, and commonsense reasoning.

    \item 
 \textbf{LLaVA-Bench (In-the-Wild)}~\cite{llava}:
    LLaVA-Wild consists of diverse real-world images paired with open-ended questions spanning various domains.
    It is designed to assess robustness and generalization under unconstrained scenarios.

     \item 
\textbf{MM-Vet}~\cite{yu2023mmvet}:
    MM-Vet evaluates a model’s ability to conduct visually grounded conversations across multiple reasoning skills.
    It covers tasks such as object recognition, spatial reasoning, and commonsense inference. 
\end{itemize}

\end{document}